\documentclass[10p,letterpaper]{article}

\usepackage[preprint]{neurips_2023}    

\usepackage{graphicx}
\usepackage{amsmath}
\usepackage{amssymb}
\usepackage{booktabs}
\usepackage[affil-it]{authblk}
\usepackage[table]{xcolor} 
\usepackage{multirow}

\setcitestyle{square,comma,numbers,sort&compress}
\usepackage[pagebackref,breaklinks,colorlinks]{hyperref}

\usepackage[capitalize]{cleveref}
\crefname{section}{Sec.}{Secs.}
\Crefname{section}{Section}{Sections}
\Crefname{table}{Table}{Tables}
\crefname{table}{Tab.}{Tabs.}

\begin{document}

\title{DME-Driver: Integrating Human Decision Logic and 3D Scene Perception in Autonomous Driving}

\author{Wencheng Han}
\author{Dongqian Guo}
\author{Cheng-Zhong Xu}
\author{Jianbing Shen\thanks{Corresponding author}}

\affil{SKL-IOTSC, CIS, University of Macau}

\affil{wencheng256@gmail.com, jianbingshen@um.edu.mo}
\maketitle

\begin{abstract}
In the field of autonomous driving, two important features of autonomous driving car systems are the explainability of decision logic and the accuracy of environmental perception. This paper introduces DME-Driver, a new autonomous driving system that enhances the performance and reliability of autonomous driving system. DME-Driver utilizes a powerful vision language model as the decision-maker and a planning-oriented perception model as the control signal generator. To ensure explainable and reliable driving decisions, the logical decision-maker is constructed based on a large vision language model. This model follows the logic employed by experienced human drivers and makes decisions in a similar manner. On the other hand, the generation of accurate control signals relies on precise and detailed environmental perception, which is where 3D scene perception models excel. Therefore, a planning oriented perception model is employed as the signal generator. It translates the logical decisions made by the decision-maker into accurate control signals for the self-driving cars. To effectively train the proposed model, a new dataset for autonomous driving was created. This dataset encompasses a diverse range of human driver behaviors and their underlying motivations. By leveraging this dataset, our model achieves high-precision planning accuracy through a logical thinking process.
\end{abstract}

\section{Introduction}
\label{sec:intro}

\begin{figure}
  \centering
  \includegraphics[width = 0.7 \linewidth]{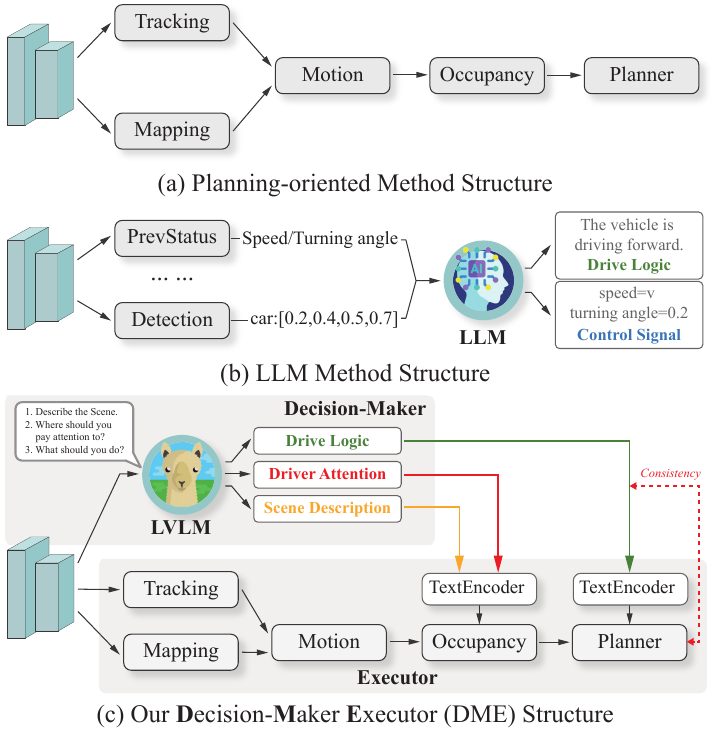}
  \caption{\textbf{Comparative Structures in Autonomous Driving Systems:} (a) depicts an planning-oriented autonomous driving system that optimizes for overall performance using planning results, but lacks interpretability; (b) shows an LLM-based autonomous driving system capable of producing reasonable control signals, yet unable to fully leverage perception tasks; 
  (c) illustrates our DME-Driver Autonomous Driving System, which effectively strengths both planning-oriented~\cite{hu2023planning} and LVLM~\cite{liu2023improved} models. 
  }
  \label{Fig:motivation}
\vspace{-4mm}
\end{figure} 

Autonomous driving systems represent a significant advancement in automotive cars technology, combining computer vision and artificial intelligence. These systems enable vehicles to perceive their surroundings~\cite{chen2015deepdriving, nguyen2011stereo, radecki2016all, sun2020scalability, yakovlev2009synergy}, make informed decisions~\cite{barnes2017find, fuji2014trajectory, hang2020human, isele2018navigating, schwarting2018planning}, and navigate without human intervention~\cite{hubschneider2017adding, kamath2023new, luo2019localization, shao2023safety, wu2022trajectory}. Key to these systems are sophisticated perception mechanisms and decision-making algorithms. These components allow vehicles to accurately understand their environment and make autonomous decisions, ensuring safe and efficient navigation. However, the complexity of these systems requires a focus on interpretability. Achieving interpretability is not just a technical challenge, but also a crucial step towards wider acceptance and integration of autonomous vehicles in society.

Recently, deep learning-based methods have achieved remarkable success in the realm of autonomous driving~\cite{chi2017deep, feng2018towards, huval2015empirical, liu2017visualization, maqueda2018event}. Some works~\cite{chen2023end, hu2022goal, gu2023vip3d, huang2023differentiable} proposed  planning-oriented autonomous driving systems that can be trained end-to-end. As illustrated in Fig.~\ref{Fig:motivation}(a), this system~\cite{chen2023end} encompasses several critical perception modules, including tracking, mapping, motion, and occupancy detection. The outputs from these modules are fed into a planner, which then generates the control signals for the vehicle. This approach leverages the full potential of perception models in autonomous driving cars, significantly enhancing the overall accuracy of the planning process.
However, the end-to-end nature of such methods leads to a lack of comprehensibility in terms of human driving logic, resulting in a system that lacks interpretability. When confronted with scenarios that the model fails to resolve – what we might term "bad cases" – it becomes challenging to understand the flawed decision-making process of the system. This obscurity poses a significant hurdle in troubleshooting and refining the autonomous driving system, as there is little insight into why and how these errors are made. 

Benefiting from the advancements in large language models (LLMs), recent approaches have attempted to enhance the interpretability of autonomous driving systems by integrating LLMs into their frameworks~\cite{cheng2023language, kim2020advisable, liu2023vlpd, peng2023openscene, sriram2019talk}, as shown in Fig.~\ref{Fig:motivation}(b). Leveraging their formidable logic reasoning capabilities and generalization, LLMs can effectively understand the behavioral logic of human drivers in various driving scenarios. In the face of corner cases, they can attempt to mimic human driver behavior, making reasoned judgments and decisions. Moreover, even when decision-making errors occur, the logical reasoning process of these models can be used as evidence to pinpoint the causes of erroneous judgments and to seek solutions.
However, these methods place LLMs at the core of the autonomous driving system, treating the outputs of perception modules as given conditions. This approach does not fully utilize the diverse perception tasks essential for accurate environmental perception. When perception models themselves produce erroneous predictions, these mistakes can accumulate in the decision-making process. The losses incurred from decision-making errors are then unable to retroactively optimize the perception tasks, preventing the achievement of optimal overall performance.

In this paper, we propose the DME (\textbf{D}ecision-\textbf{M}aker \textbf{E}xecutor)-Driver autonomous driving system, which synergizes the advantages of both LLMs and planning-oriented perception models. Our system harnesses the LLMs’ ability to understand human driver behavior logic while simultaneously leveraging the precise environmental perception capabilities of planning-oriented autonomous driving models. As depicted in Fig.~\ref{Fig:motivation}(c), our system comprises two key roles: the Decision-Maker and the Executor.
The Decision-Maker is based on a Large Vision Language Model (LVLM), trained through imitation learning on extensive real-world driving data and human driver behavior logic. This enables the Decision-Maker to thoroughly grasp the relationship between driving scenarios and the underlying behavioral logic. When the vehicle encounters a new scene, the Decision-Maker can simulate human-like logical assessments of key elements in the scene, determining whether to accelerate, brake, or change lanes. Additionally, the Decision-Maker can mimic a human driver's ability to describe key aspects of driving scenes and gaze information, providing the perception model with a reliable set of prior information. This helps the perception model to focus on elements of particular importance in the current scenario.

The Executor, on the other hand, is responsible for converting the Decision-Maker's instructions into precise vehicle control signals. It acts as the executor, ensuring that the high-level logic and reasoning from the Decision-Maker are effectively translated into real-world driving actions. By doing so, the Executor bridges the gap between decision-making and vehicle control, enabling the system to navigate safely and efficiently in diverse driving conditions. By fully considering these two aspects, DME-Driver aims to achieve a more detailed and human-like understanding of driving scenarios, leading to safer and more efficient autonomous driving decisions.

In summary, this paper presents four key contributions:

\begin{itemize}
    \item \textbf{DME-Driver Autonomous Driving System:}  We present the DME-Driver system, which combines the strengths of LLMs in logical reasoning and interpretability with the precise environmental sensing of planning-oriented models, improving decision-making robustness and interpretability in autonomous driving.

    \item \textbf{Human-Driver Behavior and Decision-Making (HBD) Dataset} Leveraging both open-source datasets and newly collected data, we developed a distinctive dataset that integrates human driver behavior logic with detailed environmental perception, specifically designed for training the DME-Driver system.

    \item \textbf{Decision-Maker Model Design:} Our Decision-Maker model, based on LVLM, is capable of imitating human driver instructions and focusing on important elements in the environment, providing human-like insights for better decision-making.

    \item \textbf{Executor Model Formulation:} The Executor model accurately processes environmental data and translates the Decision-Maker's instructions into accurate vehicle control signals, ensuring effective and context-aware responses in various driving situations.

\end{itemize}

Empirical evaluation demonstrate that our method achieves state-of-the-art accuracy in autonomous driving planning, significantly enhancing the system's interpretability. Every driving decision made by the system can be traced back through logs to understand the underlying driving logic, providing a level of transparency and explainability that is unprecedented in autonomous driving systems.

\section{Related Work}
\subsection{Autonomous Driving System}
The development of autonomous driving technology has evolved from traditional rule-based systems to sophisticated learning-based approaches. Initially, autonomous vehicles were governed by algorithms that heavily relied on sensor-based data and predefined rules \cite{thrun2006stanley, montremerlo2008stanford}. While these methods provided reliable results in controlled environments, they struggled with the unpredictability and complexity inherent in real-world driving conditions.

The advent of deep learning revolutionized this landscape, enabling systems to learn from vast and varied datasets of real driving scenarios \cite{badue2021self, grigorescu2020survey}. This shift to learning-based approaches has endowed autonomous systems with the flexibility and adaptability needed to navigate complex and dynamic environments more effectively, paving the way for more robust and versatile driving systems.

A significant trend in recent years are the development of end-to-end autonomous driving systems. These systems utilize deep neural networks to process sensory inputs directly into driving actions, seeking to streamline the autonomous driving process \cite{bojarski2016end}. While this approach simplifies the system architecture by eliminating modular decomposition, it raises challenges in terms of interpretability and robustness. The black box nature of these systems often hinders their ability to explain decisions and adapt to novel situations, which is critical for ensuring safety and gaining user trust in real-world applications.

\subsection{Large Language Model for Autonomous Driving}
%
The field of Large Language Models (LLMs) has seen remarkable growth, significantly advancing capabilities in natural language understanding and generation. Early models like BERT \cite{devlin2018bert} and GPT \cite{radford2019language} laid the groundwork for more complex systems such as GPT-3 \cite{brown2020language}. These models have not only excelled in generating coherent text but have also shown proficiency in understanding context and subtleties in language, making them invaluable for diverse applications.
A noteworthy advancement in LLMs is their integration with other data modalities, particularly visual data. Models like CLIP \cite{radford2021learning} and DALL-E \cite{ramesh2021zero} have demonstrated the effectiveness of combining textual and visual information, enabling a holistic understanding of multimodal content. This integration has been pivotal in broadening the applicability of LLMs to fields where context and nuance across different data types are essential.
However, LLMs are not without their challenges. The computational requirements for training and running these models are substantial, raising concerns about environmental sustainability and the digital divide \cite{strubell2019energy}. Furthermore, ethical considerations, particularly regarding biases in model outputs and their implications, are an ongoing area of concern and active research \cite{bender2021dangers}.

For a long time, autonomous driving systems have been treated as black boxes with a lack of interpretability, making it difficult to understand how decisions are made. The development of LLMs promises to solve this problem. GPT-Driver \cite{mao2023gpt} transforms the GPT-3.5 model into a motion planner for autonomous driving, which demonstrates the motion planning abilities of the LLMs. DriveGPT4 \cite{xu2023drivegpt4}, an interpretable end-to-end autonomous driving system based on LLMs, utilizes multimodal data such as videos, texts, and historical control signals. It generates textual responses to questions and predicts control signals for vehicle operation. Also, the reasoning abilities of LLMs improve the performance in perception and understanding tasks. LLM-AD \cite{elhafsi2023semantic}, a semantic anomaly detection framework utilizing LLMs' reasoning abilities, demonstrates that the LLM-based monitor aligns with human intuition in both fully end-to-end policies and classical autonomy stacks utilizing learned perception.

\section{Human Driver Behavior and Decision-Making Dataset}
\begin{figure}[h]
\centering
\hsize=\textwidth
    \includegraphics[width = \linewidth]{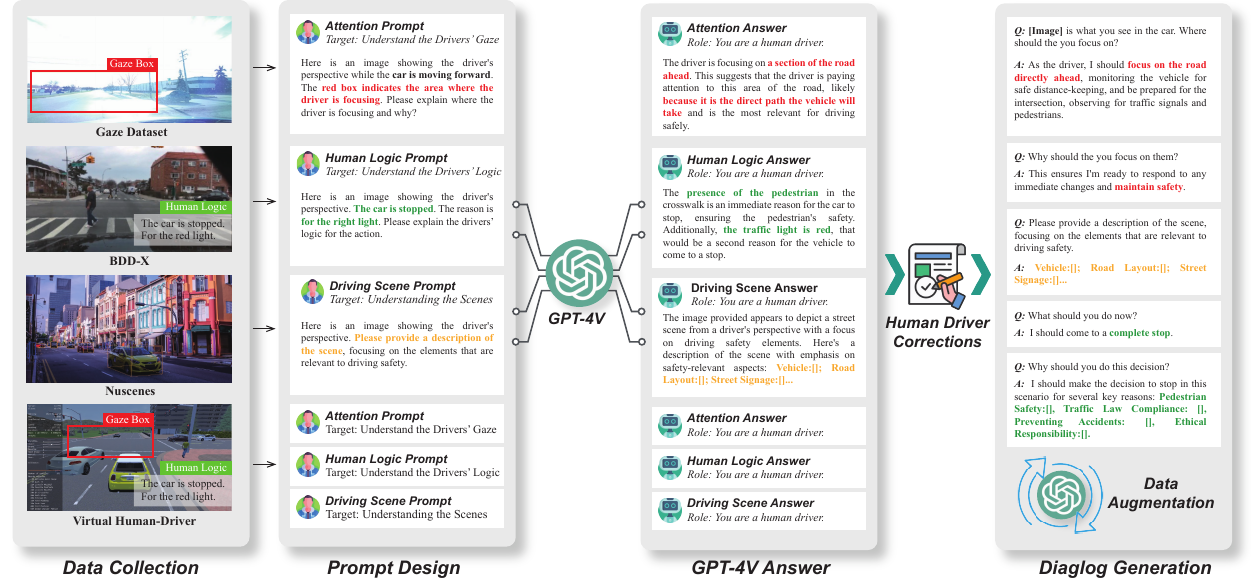}
    \caption{\textbf{HBD Dataset Annotation Pipeline:} There are four key steps in our dataset creation - Data Collection, Prompt Design, Manual Corrections, and Dialogue Generation, streamlining the process from raw data gathering to structured dialogue formation.}
    \label{fig:data}
\end{figure}

\subsection{Dataset Introduction}
In order to thoroughly investigate the relationship between the behavioral logic of human drivers and the resulting driving signals, our study focuses on four key aspects. These aspects are essential for enhancing the robustness of an autonomous driving system by understanding and mimicking human-like driving behaviors. 

\noindent\textbf{Human Driver Gaze in Driving.} Driving on roads requires real-time responses, often based on subconscious reflexes rather than deliberate reasoning. To comprehend these reflexive actions, we consider human gaze as an informative behavioral signal. During driving, human drivers instinctively focus on the most critical parts of the scene, which are typically directly influential in the driving logic of the current scenario. These elements could include traffic signals or other entities that might interact with the vehicle shortly. Understanding this gaze behavior is crucial for our system to recognize and prioritize important aspects of the driving environment.

\noindent\textbf{Human Driver's Understanding of Driving Scenes.} The way human drivers logically describe driving scenes provides a rich, purposeful understanding of the scenario. Unlike standard image captions that offer a global view, human drivers focus on elements and their interrelations that could impact driving decisions. These logical descriptions enable even other human drivers to make accurate judgments about appropriate driving actions. For instance, a description might detail \textit{a scenario at an intersection where the vehicle is in a left-turn lane with a red light for turning.}

\noindent\textbf{Decision-Making and Rationale of Human Drivers.} The decision-making process of human drivers is logical and information-rich. It encompasses the driver's synthesis of various factors to determine the appropriate action for a given scenario, along with the underlying thought process. By comprehending and emulating this aspect, an autonomous driving system can mimic human-like decision-making logic. This capability is particularly valuable in novel or challenging scenarios not encountered during training, enabling the system to make safe and reliable judgments.

\noindent\textbf{Precise Control Signals.} The ultimate output of an autonomous driving system should be structured control signals directly applicable to the vehicle. Regardless of how correct and interpretable the natural language-based driving logic is, it must be translated into concrete control commands. This translation is vital to ensure that the detailed understanding and decisions derived from human driving logic are effectively and safely executed by the autonomous vehicle.

Given the requirement for four distinct types of raw data, we faced the challenge that no existing open-source dataset comprehensively covers all these aspects. To address this gap, we integrated three different open-source datasets, each contributing one or more of the needed data types. For example, the Look-Both-Ways~\cite{kasahara2022look} provides precise human driver gaze information collected via eye-tracking devices. BDD-X~\cite{bdd100k} offers manually annotated driving decisions and their underlying reasons. Nuscenes~\cite{caesar2020nuscenes} includes detailed control signals. Leveraging the powerful generalization capabilities of LVLM, we can extrapolate the knowledge learned from each dataset to the others, even in the absence of a single dataset containing all types of annotations.

Besides this, to further enhance the LVLM’s capabilities, especially in multi-turn question-answering scenarios within the same scene, we collected a new, comprehensive virtual sub-dataset named Virtual HBD. This dataset was gathered using the Carla~\cite{dosovitskiy2017carla} simulation engine, where human drivers operated simulated driving controls such as steering wheels and pedals. This process yielded visual information and control signals for the vehicle. Additionally, drivers manually annotated their specific decisions and the reasons behind them at each moment. During the driving sessions, an eye tracker was employed to capture the drivers' gaze, focusing on the objects they observed in real-time.

Thanks to this novel Virtual HBD sub-dataset, we can link all related information in a single dialogue, significantly bolstering the LVLM’s generalization ability. This dataset uniquely enables the system to understand and mimic the comprehensive decision-making process of human drivers, covering everything from gaze patterns and scene interpretation to the rationale behind specific driving decisions and their translation into control signals.

\subsection{Data Collection and Labels Generation}
To efficiently process and label this data, we employ a combination of GPT-4V(ision) pre-annotation followed by manual corrections. Fig.~\ref{fig:data} outlines the data collection pipeline for our HBD Dataset. This pipeline comprises four main steps, starting from data collection, involving Prompt Design and manual correction, to the generation of dialogues. 

\noindent\textbf{Step 1: Data Collection} As mentioned in the previous section, our data collection integrates four sub-datasets, including three open-source datasets - Look-Both-Ways~\cite{kasahara2022look}, BDD-X~\cite{bdd100k}, and Nuscenes~\cite{caesar2020nuscenes} - and one newly collected sub-dataset, \textit{Virtual HBD}. This comprehensive combination provides a rich base of raw data encompassing various aspects of human driving behavior.

\noindent\textbf{Step 2: Prompt Design} To guide GPT-4V in analyzing driving behaviors from a human driver’s perspective, we crafted unique prompts for each data type. For example, with human gaze data, the aim is for LVLM to understand which areas a human driver focuses on in a given scene and why. We first convert gaze points into an axis-aligned bounding box by taking the minimum and maximum values of x and y coordinates from the gaze points within 24 frames. This bounding box is then drawn on the image. GPT-4V processes this enhanced image to infer the driver's focus and intentions.

\noindent\textbf{Step 3: Manual Corrections} After GPT-4V generates outputs, these are manually reviewed and corrected. Completely incorrect responses are deleted and regenerated with altered prompts. For correct responses, any details that do not align with typical human driver thought processes are manually adjusted.

\noindent\textbf{Step 4: Dialogue Generation} Once all data are transformed into detailed textual information, the next step is organizing this information into dialogues, which involves three sub-steps:

\noindent \textit{First-person Conversion}: All pronouns in the dialogues are converted to the first person. This is to ensure that the subsequent LVLM Decision-Maker model can process the dialogues from the perspective of the human driver.

\noindent  \textit{Combining Multi-turn Dialogues}: The labeling process typically involves single-question prompts. For practical use, multi-turn dialogues offer more contextual clues for the LVLM. We thus concatenate different types of Q\&A information into multi-turn dialogues, as shown in Fig.~\ref{fig:data}. The number of dialogue turns varies – one to three turns for open-source data, depending on the information available, and up to five turns for the Virtual HBD data.

\noindent  \textit{Data Augmentation}: To diversify the question-answer labels and enhance the model's generalization capabilities, we utilize GPT-3.5 to rewrite the generated dialogues. This process involves changing the form of the dialogues while keeping their content consistent.

\section{The proposed DME-Driver Autonomous Driving System}

\begin{figure}[h]
\centering
\hsize=\textwidth
    \includegraphics[width = \linewidth]{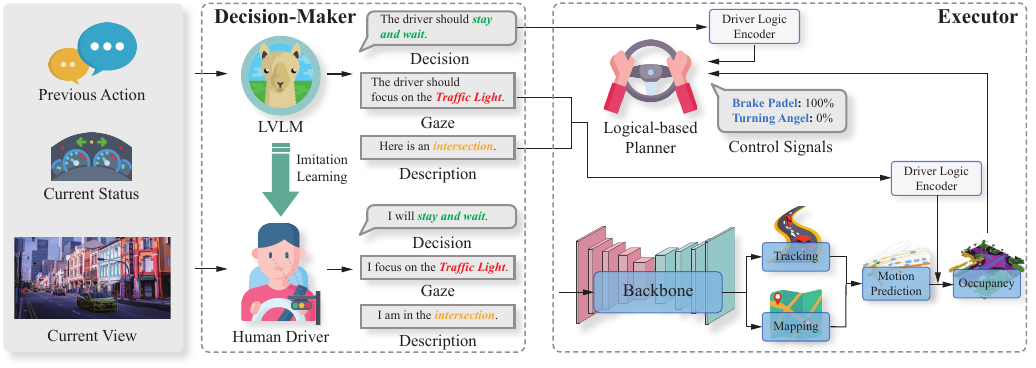}
    \vspace{-1.5mm}
    \caption{\textbf{The Detailed
    DME-Driver Autonomous Driving System:} There are two main components in our system. The \textbf{Decision-Maker} serves as the decision-maker, emulating human drivers' logic in various driving scenarios. The \textbf{Executor} then translates the Decision-Maker's instructions into precise vehicle control signals, ensuring effective execution of driving decisions.}
    \label{fig:network}
\end{figure}

In our DME-Driver Autonomous Driving System, as illustrated in Fig.~\ref{fig:network}, the system is divided into two main components: the Decision-Maker and the Executor. The Decision-Maker acts as the central decision-maker, synthesizing vehicle status and current visual inputs to emulate a human driver's logical judgments. Its output is expressed in natural language, providing a logical and interpretable narrative of driving decisions. This feature is particularly valuable for diagnosing and understanding 'bad cases' in driving scenarios, as these natural language logs offer insights into the causes of incorrect decisions. However, as natural language cannot directly control a vehicle, the system incorporates the Executor network, functioning as a translator. This network converts the Decision-Maker's linguistic outputs into precise vehicle control commands. The detailed architecture and functions of both the Decision-Maker and the Executor networks are crucial to the system's effectiveness and will be elaborated in Sections 4.1 and 4.2 respectively. 

\subsection{Decision-Maker: Human Driver Logic Understanding}
The Decision-Maker in our DME-Driver system is a sophisticated Large Vision Language Model (LVLM) designed to simulate the decision-making process of human drivers. In our experiments, we utilize LLaVA~\cite{liu2023improved} as the baseline network for the Decision-Maker. This component is engineered to process inputs from three different modalities: visual inputs from the current and previous scenes, textual inputs in the form of prompts, and current status information detailing the vehicle's operating state. 

\noindent\textbf{Visual Input:} To process the visual input, we utilize a pretrained CLIP~\cite{radford2021learning} visual encoder $E_{CLIP}$. This encoder converts the visual information into feature tokens. To better comprehend the context of driving scenes, we enhance the input by concatenating the previous three key frames with the current frames into an image array:
\begin{equation}
    \text{F}_v = E_{CLIP}(F_{t-3} \oplus F_{t-2} \oplus F_{t-1} \oplus F_{t})
\end{equation}

\noindent\textbf{Textual and Status Input:} The prompt inputs $T_p$ and current status $T_s$ information are handled using a methodology similar to RT-2~\cite{brohan2023rt}, where a text tokenizer is employed to encode these inputs uniformly.  After encoding, the tokens representing both visual and textual information are concatenated and fed into the LLaMA 2~\cite{touvron2023llama} model for processing:
\begin{equation}
    \begin{aligned}
    \text{F}_t &= \text{Tokenizer}(T_p) \oplus \text{Tokenizer}(T_s) \\
    \text{t} &= \text{LLaMA\_2}(\text{F}_t + \text{F}_v)
    \end{aligned}
\end{equation}
This integrated approach allows the Decision-Maker to consider all aspects of the driving scenario, ensuring a comprehensive understanding and simulation of human-like decision-making processes. The final step involves a de-tokenizer, which maps the output tokens back into natural language.


\subsection{Executor: The Control Signal Generator}
As depicted in Fig.~\ref{fig:network}, our Executor network in the DME-Driver system is designed based on the UniAD~\cite{hu2023planning} planning-oriented autonomous driving framework, featuring 4 distinct components:

\noindent\textbf{Backbone Network:} The initial layer of the Executor network is a backbone network, which is responsible for extracting features from multi-view vision inputs. This network forms the foundation for subsequent feature processing and interpretation. Following the backbone network, the extracted image features are transformed into Bird's Eye View (BEV) features through a process similar to BEVFormer.

\noindent\textbf{Perception Modules:} The next stage consists of four specialized perception modules:

\noindent\textit{TrackFormer} is designed for detecting and tracking various elements within the driving scene.
        
\noindent\textit{MapFormer} generates a segmented map in BEV, providing detailed spatial information about the environment.

\noindent\textit{MotionFormer} predicts the motion trajectories of each element within the scene.

\noindent\textit{OccFormer} is responsible for generating occupancy information, indicating the areas within the scene that are occupied and those that are free.

\noindent\textbf{Planning Module:} Following the perception modules, the Planning Module takes the output tokens from these modules as its input. This module's primary function is to generate the predicted control signals for the vehicle.

\noindent\textbf{Driver Logic Encoder:} Distinct from the UniAD system, our Executor network incorporates additional enhancements for the OccFormer and the Planning module. The OccFormer combines textual information from scene descriptions and gaze data, while the Planning module integrates decision-related text. Specifically, we've integrated a Bert-based text encoder $E_{bert}$ to process corresponding textual inputs:
\begin{equation}
\begin{aligned}
    &T_{occ} = E_{bert}(t_{gaze}) \oplus E_{bert}(t_{description}) \\
    &T_{planner} = E_{bert}(t_{decision}),
\end{aligned}
\end{equation}
where $T_{occ}$ represents the text encoding for the OccFormer, while $T_{planner}$ represents the text encoding for the Planning module. $t_{gaze}$, $t_{description}$, and $t_{decision}$ represent the answers generated by the Decision-Maker for the Gaze, Scene Description, and Decision Making questions, respectively. After encoding the text, we combine the BEV feature $B$ generated by the backbone network with the corresponding text encoding using a transformer fusion structure named LogicalFusioner. In this structure, we consider the BEV feature as the query and the text encoding as the key and value. After the aggregating of the multi-head attention, we add a shortcut connection to the original $B$ and produce the enhanced BEV feature $B'$:
\begin{equation}
\begin{aligned}
    B' &= \text{LogicalFusioner}(B, T)  \\
       &= \text{MHA}(Q=B, K=T, V=T) + B
\end{aligned}
\end{equation}
The inclusion of the TextEncoder in the Executor network enables it to go beyond just processing visual information; it allows for the integration of decision-making information and scene understanding provided by the Decision-Maker, emulating human driver insights for more comprehensive and context-aware driving decisions.

\subsection{Training}
The training of the DME-Driver system is streamlined into two essential steps. Firstly, training the Decision-Maker involves using multi-type human driver decision data to understand the human driving logic. Secondly, the training focuses on the Executor, which is trained using Decision-Maker instructions, perception labels, and control signals. By utilizing this data, the Executor can learn how to accurately transform instructions into control signals.

\noindent \textbf{Decision-Maker Training:}
The training of the Decision-Maker network in our DME-Driver system encompasses two critical stages: pretraining and fine-tuning. Initially, the model undergoes pretraining on diverse datasets, including 593K image-text pairs from CC3M~\cite{changpinyo2021conceptual} and 100K video-text pairs from WebVid-10M~\cite{bain2021frozen}, focusing on general video-text alignment. This phase involves training the video tokenizer while keeping the CLIP encoder and LLM weights fixed. The fine-tuning stage then tailors the model to the specific needs of interpretable autonomous driving. Here, the LLM is trained alongside the visual tokenizer using 30K video-text pairs, from the proposed HBD Dataset, and supplemented with 80K instruction-following image-text pairs from LLaVA~\cite{liu2023improved}. 

\noindent \textbf{Executor Training:}
The training of the Executor component in our DME-Driver system primarily follows the setup utilized by UniAD~\cite{hu2023planning}. However, we introduce specific modifications to enhance the system's consistency. Initially, similar to UniAD, we start by jointly training the perception parts, namely the tracking and mapping modules, for six epochs. We then proceed to an end-to-end training phase, which lasts for 20 epochs and encompasses all perception, prediction, and planning modules. To ensure alignment between the output signals of the planning module and the decisions made by the Decision-Maker, we introduce an additional reinforcement learning component during the training of the planning module. This component applies a penalty whenever the control signals deviate from the Decision-Maker's decisions. Specifically, we category the Decision-Maker's decisions into eight distinct types, such as moving forward, turning left, turning right, among others. For each of these decision types, we've established specific rules to determine whether a given control signal corresponds to one of these categories.

\section{Experiment}

\subsection{Human Driver Logic}
To assess the accuracy of the Decision-Maker in mimicking human driver decision-making and judgment in driving scenarios, we conducted an evaluation using the test set of the HBD dataset. Our primary goal was to determine whether the Decision-Maker could accurately replicate the focus areas, scene descriptions, reasoning, and decisions characteristic of human drivers.

In our evaluation process, we adapted a method similar to that used in DriveGPT4, utilizing an advanced version of ChatGPT to generate assessment metrics. Leveraging ChatGPT’s advanced reasoning capabilities, we designed it to provide a numerical score ranging from 0 to 1 for each prediction, where a higher score indicates better accuracy in mirroring human-like decision-making. This scoring method allows for a nuanced and comprehensive assessment of the Decision-Maker's performance.

The evaluation of the Decision-Maker's accuracy was conducted across four key dimensions:

\noindent\textbf{Gaze:} Assessing the accuracy of the Decision-Maker in identifying areas of focus during the driving process.

\noindent\textbf{Scene Understanding:} Evaluating how precisely the Decision-Maker describes the elements present in the current driving scene.

\noindent\textbf{Reasoning:} Analyzing the correctness of the logic employed by the Decision-Maker in making driving decisions.

\noindent\textbf{Decision:} Determining the accuracy of the final driving decisions made by the Decision-Maker.

To benchmark our system's performance, we compared the Decision-Maker's accuracy with that of other general-purpose large models, including LLaVA and GPT-4V, in similar scenarios. The outcomes of these comparisons are detailed in Table~\ref{table:planning}.

\begin{table}[h!]
  \centering
    \caption{\textbf{Comparative Analysis of Driving Logic Understanding:} This table contrasts our DME-Driver system with other large language models, emphasizing the proficiency in comprehending and interpreting driving logic.}
    \label{table:planning}
  \begin{tabular}{lccc}
    \toprule
    \multirow{2}{*}{Method}& \multicolumn{1}{c}{Gaze} & \multicolumn{1}{c}{Scene Understanding} & \multicolumn{1}{c}{Logic} \\
    \cmidrule(lr){2-2} \cmidrule(lr){3-3} \cmidrule(lr){4-4}
     & ChatGPT4$\uparrow$ & ChatGPT4$\uparrow$ & ChatGPT4$\uparrow$ \\
    \midrule
    LLaVA-7B\cite{liu2023improved} & 45.3 & 60.1 & 45.7 \\
    LLaVA-13B\cite{liu2023improved} & 48.3 & 65.2 & 48.2 \\
    GPT-4V\cite{yang2023dawn} & 75.3 & 79.2 & 65.4 \\
    \rowcolor{gray!25} 
    DME-Driver & 85.2 & 86.5 & 80.3 \\
    \bottomrule
  \end{tabular}
\end{table}

\subsection{Planning}
The aim of this experiment is to validate the accuracy of the entire DME-Driver system in making autonomous driving decisions. Following methodologies established in previous works, we focus on evaluating the final decision accuracy of the system.

As shown in Table~\ref{table:logic}, the results of our experiments demonstrate that the DME-Driver system successfully harnesses the logic-driven prompts from the Decision-Maker, enhancing the decision-making precision of the planning module. This integration not only leads to higher decision accuracy in diverse driving situations but also maintains a detailed log of the decision-making process. The ability to trace back and understand the rationale behind each decision is a critical aspect of our system, adding a layer of interpretability and accountability that is often lacking in autonomous driving systems.

\begin{table}[ht]
\centering
\caption{\textbf{Comparison of Planning Accuracy:} This table showcases a comparative analysis between our DME-Driver system and state-of-the-art methods, highlighting the advancements in planning accuracy achieved by our approach.}
\label{table:logic}
\begin{tabular}{lccccc|cccc}
\toprule
\multirow{2}{*}{Method} & \multirow{2}{*}{Input} & \multicolumn{4}{c}{L2(m)$\downarrow$} & \multicolumn{4}{c}{Col. Rate(\%)$\downarrow$} \\
\cmidrule(lr){3-6} \cmidrule(lr){7-10}
& & 1s & 2s & 3s & Avg. & 1s & 2s & 3s & Avg. \\
\midrule
NMP\cite{zeng2019end} & Lidar & - & - & 2.31 & - & - & - & 1.92 & - \\
SA-NMP\cite{zeng2019end} & Lidar & - & - & 2.05 & - & - & - & 1.59 & - \\
FF\cite{hu2021safe} & Lidar & 0.55 & 1.20 & 2.54 & 1.43 & 0.06 & 0.17 & 1.07 & 0.43 \\
EO\cite{khurana2022differentiable} & Lidar & 0.67 & 1.36 & 2.78 & 1.60 & 0.04 & 0.09 & 0.88 & 0.33 \\
\midrule
ST-P3\cite{hu2022st} & Vision & 1.33 & 2.11 & 2.90 & 2.11 & 0.23 & 0.62 & 1.27 & 0.71 \\
UniAD\cite{hu2023planning} & Vision & 0.48 & 0.96 & 1.65 & 1.03 & 0.05 & 0.17 & 0.71 & 0.31 \\
\midrule
\rowcolor{gray!25}DME-Driver & Vision & 0.45 & 0.91 & 1.58 & 0.98 & 0.05 & 0.15 & 0.68 & 0.29 \\
\bottomrule
\end{tabular}
\end{table}

\begin{table}[h!]
  \centering
  \caption{\textbf{Ablation Study Results:} This table presents the impact of various components within our DME-Driver system, illustrating how each part contributes to the overall effectiveness and decision-making accuracy.}
  \label{table:ablation}
  \begin{tabular}{lcc}
    \toprule
    Module & L2(m)$\downarrow$ & Col.Rate(\%)$\downarrow$ \\
    \midrule
    Executor & 1.03 & 0.31 \\
    GT+Executor & 0.94 & 0.28 \\
    Decision-Maker + Executor & 0.96 & 0.28 \\
    \rowcolor{gray!25}
    Decision-Maker + Executor + CL & 0.98 & 0.29 \\
    \bottomrule
  \end{tabular}
\end{table}

\subsection{Ablation Study}
In our ablation study of the DME-Driver system, we methodically dissected the impact of each component on decision-making effectiveness, as shown in Table~\ref{table:ablation}. We began by assessing the standalone performance of the Executor without Decision-Maker guidance, establishing a baseline. Next, we evaluated the impact of substituting the Decision-Maker's guidance with ground truth language cues, observing potential improvements. Following this, we examined the combined performance of the Decision-Maker and Executor, gauging their collaborative efficiency. A crucial addition was the implementation of a consistency loss mechanism, slightly reducing performance metrics but significantly enhancing decision alignment between Executor and Decision-Maker. 

\section{Conclusion}
In addressing the challenges of interpretability and insufficient use of human driver behavior patterns in autonomous driving systems, this paper introduces the DME-Driver Autonomous Driving System, a novel framework comprising two integral components: the Decision-Maker and the Executor. The Decision-Maker serves as the central decision-maker, adeptly understanding and emulating human driver logic, thus ensuring each action taken by the system is both logical and accountable. The Executor complements this by effectively translating the nuanced decisions of the Decision-Maker into precise vehicle control signals, harnessing the strengths of perception tasks and planning algorithms. To facilitate comprehensive training and understanding of human driver behavior, we developed the HBD dataset, rich in diverse and essential driving information such as gaze, decision logic, and operational signals. Our empirical tests showcase the system’s capability to accurately replicate human driver reasoning and actions. Combined with the Decision-Maker's guidance, the Executor successfully converts these into operational commands, elevating the overall decision-making efficacy to a state-of-the-art level. This achievement not only demonstrates the effectiveness of the DME-Driver system but also marks a significant leap forward in the field of autonomous driving technology.

{\small

\bibliography{egbib}

\begin{thebibliography}{59}
\providecommand{\natexlab}[1]{#1}
\providecommand{\url}[1]{\texttt{#1}}
\expandafter\ifx\csname urlstyle\endcsname\relax
  \providecommand{\doi}[1]{doi: #1}\else
  \providecommand{\doi}{doi: \begingroup \urlstyle{rm}\Url}\fi

\bibitem[Badue et~al.(2021)Badue, Guidolini, Carneiro, Azevedo, Cardoso, Forechi, Jesus, Berriel, Paixao, Mutz, et~al.]{badue2021self}
Claudine Badue, R{\^a}nik Guidolini, Raphael~Vivacqua Carneiro, Pedro Azevedo, Vinicius~B Cardoso, Avelino Forechi, Luan Jesus, Rodrigo Berriel, Thiago~M Paixao, Filipe Mutz, et~al.
\newblock Self-driving cars: A survey.
\newblock \emph{Expert Systems with Applications}, 165:\penalty0 113816, 2021.

\bibitem[Bain et~al.(2021)Bain, Nagrani, Varol, and Zisserman]{bain2021frozen}
Max Bain, Arsha Nagrani, G{\"u}l Varol, and Andrew Zisserman.
\newblock Frozen in time: A joint video and image encoder for end-to-end retrieval.
\newblock In \emph{Proceedings of the IEEE/CVF International Conference on Computer Vision}, pages 1728--1738, 2021.

\bibitem[Barnes et~al.(2017)Barnes, Maddern, and Posner]{barnes2017find}
Dan Barnes, Will Maddern, and Ingmar Posner.
\newblock Find your own way: Weakly-supervised segmentation of path proposals for urban autonomy.
\newblock In \emph{2017 IEEE International Conference on Robotics and Automation (ICRA)}, pages 203--210. IEEE, 2017.

\bibitem[Bender et~al.(2021)Bender, Gebru, McMillan-Major, and Shmitchell]{bender2021dangers}
Emily~M Bender, Timnit Gebru, Angelina McMillan-Major, and Shmargaret Shmitchell.
\newblock On the dangers of stochastic parrots: Can language models be too big?
\newblock In \emph{Proceedings of the 2021 ACM conference on fairness, accountability, and transparency}, pages 610--623, 2021.

\bibitem[Bojarski et~al.(2016)Bojarski, Del~Testa, Dworakowski, Firner, Flepp, Goyal, Jackel, Monfort, Muller, Zhang, et~al.]{bojarski2016end}
Mariusz Bojarski, Davide Del~Testa, Daniel Dworakowski, Bernhard Firner, Beat Flepp, Prasoon Goyal, Lawrence~D Jackel, Mathew Monfort, Urs Muller, Jiakai Zhang, et~al.
\newblock End to end learning for self-driving cars.
\newblock \emph{arXiv preprint arXiv:1604.07316}, 2016.

\bibitem[Brohan et~al.(2023)Brohan, Brown, Carbajal, Chebotar, Chen, Choromanski, Ding, Driess, Dubey, Finn, et~al.]{brohan2023rt}
Anthony Brohan, Noah Brown, Justice Carbajal, Yevgen Chebotar, Xi~Chen, Krzysztof Choromanski, Tianli Ding, Danny Driess, Avinava Dubey, Chelsea Finn, et~al.
\newblock Rt-2: Vision-language-action models transfer web knowledge to robotic control.
\newblock \emph{arXiv preprint arXiv:2307.15818}, 2023.

\bibitem[Brown et~al.(2020)Brown, Mann, Ryder, Subbiah, Kaplan, Dhariwal, Neelakantan, Shyam, Sastry, Askell, et~al.]{brown2020language}
Tom Brown, Benjamin Mann, Nick Ryder, Melanie Subbiah, Jared~D Kaplan, Prafulla Dhariwal, Arvind Neelakantan, Pranav Shyam, Girish Sastry, Amanda Askell, et~al.
\newblock Language models are few-shot learners.
\newblock \emph{Advances in neural information processing systems}, 33:\penalty0 1877--1901, 2020.

\bibitem[Caesar et~al.(2020)Caesar, Bankiti, Lang, Vora, Liong, Xu, Krishnan, Pan, Baldan, and Beijbom]{caesar2020nuscenes}
Holger Caesar, Varun Bankiti, Alex~H Lang, Sourabh Vora, Venice~Erin Liong, Qiang Xu, Anush Krishnan, Yu~Pan, Giancarlo Baldan, and Oscar Beijbom.
\newblock nuscenes: A multimodal dataset for autonomous driving.
\newblock In \emph{Proceedings of the IEEE/CVF conference on computer vision and pattern recognition}, pages 11621--11631, 2020.

\bibitem[Changpinyo et~al.(2021)Changpinyo, Sharma, Ding, and Soricut]{changpinyo2021conceptual}
Soravit Changpinyo, Piyush Sharma, Nan Ding, and Radu Soricut.
\newblock Conceptual 12m: Pushing web-scale image-text pre-training to recognize long-tail visual concepts.
\newblock In \emph{Proceedings of the IEEE/CVF Conference on Computer Vision and Pattern Recognition}, pages 3558--3568, 2021.

\bibitem[Chen et~al.(2015)Chen, Seff, Kornhauser, and Xiao]{chen2015deepdriving}
Chenyi Chen, Ari Seff, Alain Kornhauser, and Jianxiong Xiao.
\newblock Deepdriving: Learning affordance for direct perception in autonomous driving.
\newblock In \emph{Proceedings of the IEEE international conference on computer vision}, pages 2722--2730, 2015.

\bibitem[Chen et~al.(2023)Chen, Wu, Chitta, Jaeger, Geiger, and Li]{chen2023end}
Li~Chen, Penghao Wu, Kashyap Chitta, Bernhard Jaeger, Andreas Geiger, and Hongyang Li.
\newblock End-to-end autonomous driving: Challenges and frontiers.
\newblock \emph{arXiv preprint arXiv:2306.16927}, 2023.

\bibitem[Cheng et~al.(2023)Cheng, Yin, Li, Yang, and Shen]{cheng2023language}
Wenhao Cheng, Junbo Yin, Wei Li, Ruigang Yang, and Jianbing Shen.
\newblock Language-guided 3d object detection in point cloud for autonomous driving.
\newblock \emph{arXiv preprint arXiv:2305.15765}, 2023.

\bibitem[Chi and Mu(2017)]{chi2017deep}
Lu~Chi and Yadong Mu.
\newblock Deep steering: Learning end-to-end driving model from spatial and temporal visual cues.
\newblock \emph{arXiv preprint arXiv:1708.03798}, 2017.

\bibitem[Devlin et~al.(2018)Devlin, Chang, Lee, and Toutanova]{devlin2018bert}
Jacob Devlin, Ming-Wei Chang, Kenton Lee, and Kristina Toutanova.
\newblock Bert: Pre-training of deep bidirectional transformers for language understanding.
\newblock \emph{arXiv preprint arXiv:1810.04805}, 2018.

\bibitem[Dosovitskiy et~al.(2017)Dosovitskiy, Ros, Codevilla, Lopez, and Koltun]{dosovitskiy2017carla}
Alexey Dosovitskiy, German Ros, Felipe Codevilla, Antonio Lopez, and Vladlen Koltun.
\newblock Carla: An open urban driving simulator.
\newblock In \emph{Conference on robot learning}, pages 1--16. PMLR, 2017.

\bibitem[Elhafsi et~al.(2023)Elhafsi, Sinha, Agia, Schmerling, Nesnas, and Pavone]{elhafsi2023semantic}
Amine Elhafsi, Rohan Sinha, Christopher Agia, Edward Schmerling, Issa~AD Nesnas, and Marco Pavone.
\newblock Semantic anomaly detection with large language models.
\newblock \emph{Autonomous Robots}, pages 1--21, 2023.

\bibitem[Feng et~al.(2018)Feng, Rosenbaum, and Dietmayer]{feng2018towards}
Di~Feng, Lars Rosenbaum, and Klaus Dietmayer.
\newblock Towards safe autonomous driving: Capture uncertainty in the deep neural network for lidar 3d vehicle detection.
\newblock In \emph{2018 21st international conference on intelligent transportation systems (ITSC)}, pages 3266--3273. IEEE, 2018.

\bibitem[Fuji et~al.(2014)Fuji, Xiang, Tazaki, Levedahl, and Suzuki]{fuji2014trajectory}
Hiroshi Fuji, Jingyu Xiang, Yuichi Tazaki, Blaine Levedahl, and Tatsuya Suzuki.
\newblock Trajectory planning for automated parking using multi-resolution state roadmap considering non-holonomic constraints.
\newblock In \emph{2014 IEEE Intelligent Vehicles Symposium Proceedings}, pages 407--413. IEEE, 2014.

\bibitem[Grigorescu et~al.(2020)Grigorescu, Trasnea, Cocias, and Macesanu]{grigorescu2020survey}
Sorin Grigorescu, Bogdan Trasnea, Tiberiu Cocias, and Gigel Macesanu.
\newblock A survey of deep learning techniques for autonomous driving.
\newblock \emph{Journal of Field Robotics}, 37\penalty0 (3):\penalty0 362--386, 2020.

\bibitem[Gu et~al.(2023)Gu, Hu, Zhang, Chen, Wang, Wang, and Zhao]{gu2023vip3d}
Junru Gu, Chenxu Hu, Tianyuan Zhang, Xuanyao Chen, Yilun Wang, Yue Wang, and Hang Zhao.
\newblock Vip3d: End-to-end visual trajectory prediction via 3d agent queries.
\newblock In \emph{Proceedings of the IEEE/CVF Conference on Computer Vision and Pattern Recognition}, pages 5496--5506, 2023.

\bibitem[Hang et~al.(2020)Hang, Lv, Xing, Huang, and Hu]{hang2020human}
Peng Hang, Chen Lv, Yang Xing, Chao Huang, and Zhongxu Hu.
\newblock Human-like decision making for autonomous driving: A noncooperative game theoretic approach.
\newblock \emph{IEEE Transactions on Intelligent Transportation Systems}, 22\penalty0 (4):\penalty0 2076--2087, 2020.

\bibitem[Hu et~al.(2021)Hu, Huang, Dolan, Held, and Ramanan]{hu2021safe}
Peiyun Hu, Aaron Huang, John Dolan, David Held, and Deva Ramanan.
\newblock Safe local motion planning with self-supervised freespace forecasting.
\newblock In \emph{Proceedings of the IEEE/CVF Conference on Computer Vision and Pattern Recognition}, pages 12732--12741, 2021.

\bibitem[Hu et~al.(2022{\natexlab{a}})Hu, Chen, Wu, Li, Yan, and Tao]{hu2022st}
Shengchao Hu, Li~Chen, Penghao Wu, Hongyang Li, Junchi Yan, and Dacheng Tao.
\newblock St-p3: End-to-end vision-based autonomous driving via spatial-temporal feature learning.
\newblock In \emph{European Conference on Computer Vision}, pages 533--549. Springer, 2022{\natexlab{a}}.

\bibitem[Hu et~al.(2022{\natexlab{b}})Hu, Yang, Chen, Li, Sima, Zhu, Chai, Du, Lin, Wang, et~al.]{hu2022goal}
Yihan Hu, Jiazhi Yang, Li~Chen, Keyu Li, Chonghao Sima, Xizhou Zhu, Siqi Chai, Senyao Du, Tianwei Lin, Wenhai Wang, et~al.
\newblock Goal-oriented autonomous driving.
\newblock \emph{arXiv preprint arXiv:2212.10156}, 2022{\natexlab{b}}.

\bibitem[Hu et~al.(2023)Hu, Yang, Chen, Li, Sima, Zhu, Chai, Du, Lin, Wang, et~al.]{hu2023planning}
Yihan Hu, Jiazhi Yang, Li~Chen, Keyu Li, Chonghao Sima, Xizhou Zhu, Siqi Chai, Senyao Du, Tianwei Lin, Wenhai Wang, et~al.
\newblock Planning-oriented autonomous driving.
\newblock In \emph{Proceedings of the IEEE/CVF Conference on Computer Vision and Pattern Recognition}, pages 17853--17862, 2023.

\bibitem[Huang et~al.(2023)Huang, Liu, Wu, and Lv]{huang2023differentiable}
Zhiyu Huang, Haochen Liu, Jingda Wu, and Chen Lv.
\newblock Differentiable integrated motion prediction and planning with learnable cost function for autonomous driving.
\newblock \emph{IEEE transactions on neural networks and learning systems}, 2023.

\bibitem[Hubschneider et~al.(2017)Hubschneider, Bauer, Weber, and Z{\"o}llner]{hubschneider2017adding}
Christian Hubschneider, Andre Bauer, Michael Weber, and J~Marius Z{\"o}llner.
\newblock Adding navigation to the equation: Turning decisions for end-to-end vehicle control.
\newblock In \emph{2017 IEEE 20th international conference on intelligent transportation systems (ITSC)}, pages 1--8. IEEE, 2017.

\bibitem[Huval et~al.(2015)Huval, Wang, Tandon, Kiske, Song, Pazhayampallil, Andriluka, Rajpurkar, Migimatsu, Cheng-Yue, et~al.]{huval2015empirical}
Brody Huval, Tao Wang, Sameep Tandon, Jeff Kiske, Will Song, Joel Pazhayampallil, Mykhaylo Andriluka, Pranav Rajpurkar, Toki Migimatsu, Royce Cheng-Yue, et~al.
\newblock An empirical evaluation of deep learning on highway driving.
\newblock \emph{arXiv preprint arXiv:1504.01716}, 2015.

\bibitem[Isele et~al.(2018)Isele, Rahimi, Cosgun, Subramanian, and Fujimura]{isele2018navigating}
David Isele, Reza Rahimi, Akansel Cosgun, Kaushik Subramanian, and Kikuo Fujimura.
\newblock Navigating occluded intersections with autonomous vehicles using deep reinforcement learning.
\newblock In \emph{2018 IEEE international conference on robotics and automation (ICRA)}, pages 2034--2039. IEEE, 2018.

\bibitem[Kamath et~al.(2023)Kamath, Anderson, Wang, Koh, Ku, Waters, Yang, Baldridge, and Parekh]{kamath2023new}
Aishwarya Kamath, Peter Anderson, Su~Wang, Jing~Yu Koh, Alexander Ku, Austin Waters, Yinfei Yang, Jason Baldridge, and Zarana Parekh.
\newblock A new path: Scaling vision-and-language navigation with synthetic instructions and imitation learning.
\newblock In \emph{Proceedings of the IEEE/CVF Conference on Computer Vision and Pattern Recognition}, pages 10813--10823, 2023.

\bibitem[Kasahara et~al.(2022)Kasahara, Stent, and Park]{kasahara2022look}
Isaac Kasahara, Simon Stent, and Hyun~Soo Park.
\newblock Look both ways: Self-supervising driver gaze estimation and road scene saliency.
\newblock In \emph{European Conference on Computer Vision}, pages 126--142. Springer, 2022.

\bibitem[Khurana et~al.(2022)Khurana, Hu, Dave, Ziglar, Held, and Ramanan]{khurana2022differentiable}
Tarasha Khurana, Peiyun Hu, Achal Dave, Jason Ziglar, David Held, and Deva Ramanan.
\newblock Differentiable raycasting for self-supervised occupancy forecasting.
\newblock In \emph{European Conference on Computer Vision}, pages 353--369. Springer, 2022.

\bibitem[Kim et~al.(2020)Kim, Moon, Rohrbach, Darrell, and Canny]{kim2020advisable}
Jinkyu Kim, Suhong Moon, Anna Rohrbach, Trevor Darrell, and John Canny.
\newblock Advisable learning for self-driving vehicles by internalizing observation-to-action rules.
\newblock In \emph{Proceedings of the IEEE/CVF Conference on Computer Vision and Pattern Recognition}, pages 9661--9670, 2020.

\bibitem[Liu et~al.(2017)Liu, Taniguchi, Tanaka, Takenaka, and Bando]{liu2017visualization}
HaiLong Liu, Tadahiro Taniguchi, Yusuke Tanaka, Kazuhito Takenaka, and Takashi Bando.
\newblock Visualization of driving behavior based on hidden feature extraction by using deep learning.
\newblock \emph{IEEE Transactions on Intelligent Transportation Systems}, 18\penalty0 (9):\penalty0 2477--2489, 2017.

\bibitem[Liu et~al.(2023{\natexlab{a}})Liu, Li, Li, and Lee]{liu2023improved}
Haotian Liu, Chunyuan Li, Yuheng Li, and Yong~Jae Lee.
\newblock Improved baselines with visual instruction tuning, 2023{\natexlab{a}}.

\bibitem[Liu et~al.(2023{\natexlab{b}})Liu, Jiang, Zhu, and Yin]{liu2023vlpd}
Mengyin Liu, Jie Jiang, Chao Zhu, and Xu-Cheng Yin.
\newblock Vlpd: Context-aware pedestrian detection via vision-language semantic self-supervision.
\newblock In \emph{Proceedings of the IEEE/CVF Conference on Computer Vision and Pattern Recognition}, pages 6662--6671, 2023{\natexlab{b}}.

\bibitem[Luo et~al.(2019)Luo, Cao, Liu, and Benslimane]{luo2019localization}
Qian Luo, Yurui Cao, Jiajia Liu, and Abderrahim Benslimane.
\newblock Localization and navigation in autonomous driving: Threats and countermeasures.
\newblock \emph{IEEE Wireless Communications}, 26\penalty0 (4):\penalty0 38--45, 2019.

\bibitem[Mao et~al.(2023)Mao, Qian, Zhao, and Wang]{mao2023gpt}
Jiageng Mao, Yuxi Qian, Hang Zhao, and Yue Wang.
\newblock Gpt-driver: Learning to drive with gpt.
\newblock \emph{arXiv preprint arXiv:2310.01415}, 2023.

\bibitem[Maqueda et~al.(2018)Maqueda, Loquercio, Gallego, Garc{\'\i}a, and Scaramuzza]{maqueda2018event}
Ana~I Maqueda, Antonio Loquercio, Guillermo Gallego, Narciso Garc{\'\i}a, and Davide Scaramuzza.
\newblock Event-based vision meets deep learning on steering prediction for self-driving cars.
\newblock In \emph{Proceedings of the IEEE conference on computer vision and pattern recognition}, pages 5419--5427, 2018.

\bibitem[Montremerlo et~al.(2008)Montremerlo, Beeker, Bhat, and Dahlkamp]{montremerlo2008stanford}
M~Montremerlo, J~Beeker, S~Bhat, and H~Dahlkamp.
\newblock The stanford entry in the urban challenge.
\newblock \emph{Journal of Field Robotics}, 7\penalty0 (9):\penalty0 468--492, 2008.

\bibitem[Nguyen et~al.(2011)Nguyen, Michaelis, Al-Hamadi, Tornow, and Meinecke]{nguyen2011stereo}
Thien-Nghia Nguyen, Bernd Michaelis, Ayoub Al-Hamadi, Michael Tornow, and Marc-Michael Meinecke.
\newblock Stereo-camera-based urban environment perception using occupancy grid and object tracking.
\newblock \emph{IEEE Transactions on Intelligent Transportation Systems}, 13\penalty0 (1):\penalty0 154--165, 2011.

\bibitem[Peng et~al.(2023)Peng, Genova, Jiang, Tagliasacchi, Pollefeys, Funkhouser, et~al.]{peng2023openscene}
Songyou Peng, Kyle Genova, Chiyu Jiang, Andrea Tagliasacchi, Marc Pollefeys, Thomas Funkhouser, et~al.
\newblock Openscene: 3d scene understanding with open vocabularies.
\newblock In \emph{Proceedings of the IEEE/CVF Conference on Computer Vision and Pattern Recognition}, pages 815--824, 2023.

\bibitem[Radecki et~al.(2016)Radecki, Campbell, and Matzen]{radecki2016all}
Peter Radecki, Mark Campbell, and Kevin Matzen.
\newblock All weather perception: Joint data association, tracking, and classification for autonomous ground vehicles.
\newblock \emph{arXiv preprint arXiv:1605.02196}, 2016.

\bibitem[Radford et~al.(2019)Radford, Wu, Child, Luan, Amodei, Sutskever, et~al.]{radford2019language}
Alec Radford, Jeffrey Wu, Rewon Child, David Luan, Dario Amodei, Ilya Sutskever, et~al.
\newblock Language models are unsupervised multitask learners.
\newblock \emph{OpenAI blog}, 1\penalty0 (8):\penalty0 9, 2019.

\bibitem[Radford et~al.(2021)Radford, Kim, Hallacy, Ramesh, Goh, Agarwal, Sastry, Askell, Mishkin, Clark, et~al.]{radford2021learning}
Alec Radford, Jong~Wook Kim, Chris Hallacy, Aditya Ramesh, Gabriel Goh, Sandhini Agarwal, Girish Sastry, Amanda Askell, Pamela Mishkin, Jack Clark, et~al.
\newblock Learning transferable visual models from natural language supervision.
\newblock In \emph{International conference on machine learning}, pages 8748--8763. PMLR, 2021.

\bibitem[Ramesh et~al.(2021)Ramesh, Pavlov, Goh, Gray, Voss, Radford, Chen, and Sutskever]{ramesh2021zero}
Aditya Ramesh, Mikhail Pavlov, Gabriel Goh, Scott Gray, Chelsea Voss, Alec Radford, Mark Chen, and Ilya Sutskever.
\newblock Zero-shot text-to-image generation.
\newblock In \emph{International Conference on Machine Learning}, pages 8821--8831. PMLR, 2021.

\bibitem[Schwarting et~al.(2018)Schwarting, Alonso-Mora, and Rus]{schwarting2018planning}
Wilko Schwarting, Javier Alonso-Mora, and Daniela Rus.
\newblock Planning and decision-making for autonomous vehicles.
\newblock \emph{Annual Review of Control, Robotics, and Autonomous Systems}, 1:\penalty0 187--210, 2018.

\bibitem[Shao et~al.(2023)Shao, Wang, Chen, Li, and Liu]{shao2023safety}
Hao Shao, Letian Wang, Ruobing Chen, Hongsheng Li, and Yu~Liu.
\newblock Safety-enhanced autonomous driving using interpretable sensor fusion transformer.
\newblock In \emph{Conference on Robot Learning}, pages 726--737. PMLR, 2023.

\bibitem[Sriram et~al.(2019)Sriram, Maniar, Kalyanasundaram, Gandhi, Bhowmick, and Krishna]{sriram2019talk}
NN~Sriram, Tirth Maniar, Jayaganesh Kalyanasundaram, Vineet Gandhi, Brojeshwar Bhowmick, and K~Madhava Krishna.
\newblock Talk to the vehicle: Language conditioned autonomous navigation of self driving cars.
\newblock In \emph{2019 IEEE/RSJ International Conference on Intelligent Robots and Systems (IROS)}, pages 5284--5290. IEEE, 2019.

\bibitem[Strubell et~al.(2019)Strubell, Ganesh, and McCallum]{strubell2019energy}
Emma Strubell, Ananya Ganesh, and Andrew McCallum.
\newblock Energy and policy considerations for deep learning in nlp.
\newblock \emph{arXiv preprint arXiv:1906.02243}, 2019.

\bibitem[Sun et~al.(2020)Sun, Kretzschmar, Dotiwalla, Chouard, Patnaik, Tsui, Guo, Zhou, Chai, Caine, et~al.]{sun2020scalability}
Pei Sun, Henrik Kretzschmar, Xerxes Dotiwalla, Aurelien Chouard, Vijaysai Patnaik, Paul Tsui, James Guo, Yin Zhou, Yuning Chai, Benjamin Caine, et~al.
\newblock Scalability in perception for autonomous driving: Waymo open dataset.
\newblock In \emph{Proceedings of the IEEE/CVF conference on computer vision and pattern recognition}, pages 2446--2454, 2020.

\bibitem[Thrun et~al.(2006)Thrun, Montemerlo, Dahlkamp, Stavens, Aron, Diebel, Fong, Gale, Halpenny, Hoffmann, et~al.]{thrun2006stanley}
Sebastian Thrun, Mike Montemerlo, Hendrik Dahlkamp, David Stavens, Andrei Aron, James Diebel, Philip Fong, John Gale, Morgan Halpenny, Gabriel Hoffmann, et~al.
\newblock Stanley: The robot that won the darpa grand challenge.
\newblock \emph{Journal of field Robotics}, 23\penalty0 (9):\penalty0 661--692, 2006.

\bibitem[Touvron et~al.(2023)Touvron, Lavril, Izacard, Martinet, Lachaux, Lacroix, Rozi{\`e}re, Goyal, Hambro, Azhar, et~al.]{touvron2023llama}
Hugo Touvron, Thibaut Lavril, Gautier Izacard, Xavier Martinet, Marie-Anne Lachaux, Timoth{\'e}e Lacroix, Baptiste Rozi{\`e}re, Naman Goyal, Eric Hambro, Faisal Azhar, et~al.
\newblock Llama: Open and efficient foundation language models.
\newblock \emph{arXiv preprint arXiv:2302.13971}, 2023.

\bibitem[Wu et~al.(2022)Wu, Jia, Chen, Yan, Li, and Qiao]{wu2022trajectory}
Penghao Wu, Xiaosong Jia, Li~Chen, Junchi Yan, Hongyang Li, and Yu~Qiao.
\newblock Trajectory-guided control prediction for end-to-end autonomous driving: A simple yet strong baseline.
\newblock \emph{Advances in Neural Information Processing Systems}, 35:\penalty0 6119--6132, 2022.

\bibitem[Xu et~al.(2023)Xu, Zhang, Xie, Zhao, Guo, Wong, Li, and Zhao]{xu2023drivegpt4}
Zhenhua Xu, Yujia Zhang, Enze Xie, Zhen Zhao, Yong Guo, Kenneth~KY Wong, Zhenguo Li, and Hengshuang Zhao.
\newblock Drivegpt4: Interpretable end-to-end autonomous driving via large language model.
\newblock \emph{arXiv preprint arXiv:2310.01412}, 2023.

\bibitem[Yakovlev and Borisov(2009)]{yakovlev2009synergy}
SS~Yakovlev and Arkady~N Borisov.
\newblock A synergy of the rosenblatt perceptron and the jordan recurrence principle.
\newblock \emph{Automatic Control and Computer Sciences}, 43:\penalty0 31--39, 2009.

\bibitem[Yang et~al.(2023)Yang, Li, Lin, Wang, Lin, Liu, and Wang]{yang2023dawn}
Zhengyuan Yang, Linjie Li, Kevin Lin, Jianfeng Wang, Chung-Ching Lin, Zicheng Liu, and Lijuan Wang.
\newblock The dawn of lmms: Preliminary explorations with gpt-4v (ision).
\newblock \emph{arXiv preprint arXiv:2309.17421}, 9\penalty0 (1), 2023.

\bibitem[Yu et~al.(2020)Yu, Chen, Wang, Xian, Chen, Liu, Madhavan, and Darrell]{bdd100k}
Fisher Yu, Haofeng Chen, Xin Wang, Wenqi Xian, Yingying Chen, Fangchen Liu, Vashisht Madhavan, and Trevor Darrell.
\newblock Bdd100k: A diverse driving dataset for heterogeneous multitask learning.
\newblock In \emph{The IEEE Conference on Computer Vision and Pattern Recognition (CVPR)}, June 2020.

\bibitem[Zeng et~al.(2019)Zeng, Luo, Suo, Sadat, Yang, Casas, and Urtasun]{zeng2019end}
Wenyuan Zeng, Wenjie Luo, Simon Suo, Abbas Sadat, Bin Yang, Sergio Casas, and Raquel Urtasun.
\newblock End-to-end interpretable neural motion planner.
\newblock In \emph{Proceedings of the IEEE/CVF Conference on Computer Vision and Pattern Recognition}, pages 8660--8669, 2019.

\end{thebibliography}
}

\end{document}